\newcommand{\todo}[2][]{\textcolor{blue}{
\ifx&#1&%
  TODO:
\else
  TODO(#1):
\fi
#2}}

\newcommand{\review}[2][]{\textcolor{red}{
\ifx&#1&%
  REVIEW:
\else
  REVIEW(#1):
\fi
#2}}

\documentclass[journal]{IEEEtai}

\usepackage[colorlinks,urlcolor=blue,linkcolor=blue,citecolor=blue]{hyperref}

\usepackage{color,array}
\usepackage{amsfonts}
\usepackage{graphicx}
\usepackage{amsmath} 
\usepackage{pifont}
\usepackage{xcolor}
\usepackage{dsfont}
\usepackage{multirow}
\usepackage{booktabs}
\definecolor{myblue}{RGB}{28,226,255}
\usepackage[numbers,sort&compress]{natbib}

\newcommand{\firstletter}[2][]{\textcolor{myblue}{
\ifx&#1&%

\else

\fi
#2}}

\begin{document}

\title{Ensemble Learning for Large Language Models\\ in Text and Code Generation: A Survey}

\author{Mari Ashiga, Wei Jie, Fan Wu, Vardan Voskanyan, Fateme Dinmohammadi, Paul Brookes, Jingzhi Gong, and Zheng Wang
\thanks{This work was supported by the Innovate UK Knowledge Transfer Partnership (KTP) grant with KTP Reference 13804 and Project Number 10082602 (Corresponding author: Wei Jie) and submitted on December 2024 for review. }
\thanks{Mari Ashiga, Wei Jie, and Fateme Dinmohammadi are with the School of Computing and Engineering, University of West London, London W5 5RF, United Kingdom (email: mari.ashiga@uwl.ac.uk; wei.jie@uwl.ac.uk; fateme.dinmohammadi@uwl.ac.uk).}
\thanks{Fan Wu, Vardan Voskanyan, and Paul Brookes are with Turing Intelligence Technology Limited, London EC2M 2PF, United Kingdom (email: fan@turintech.ai;  vardan@turintech.ai; paul@turintech.ai).}
\thanks{Jingzhi Gong, and Zheng Wang are with the School of Computer Science, University of Leeds, Leeds LS2 9JT, United Kingdom (email: j.gong@leeds.ac.uk; z.wang5@leeds.ac.uk).}
\thanks{This paragraph will include the Associate Editor who handled your paper.}}

\markboth{Journal of IEEE Transactions on Artificial Intelligence, Vol. 00, No. 0, Month 2020}
{First A. Author \MakeLowercase{\textit{et al.}}: Bare Demo of IEEEtai.cls for IEEE Journals of IEEE Transactions on Artificial Intelligence}

\maketitle

\begin{abstract}
Generative Pretrained Transformers (GPTs) are foundational Large Language Models (LLMs) for text generation. However, individual LLMs often produce inconsistent outputs and exhibit biases, limiting their representation of diverse language patterns. The closed-source nature of many powerful LLMs further restricts industry applications due to data privacy concerns. Inspired by successes in text generation, LLM ensemble techniques are now increasingly explored for code generation. This article reviews these emerging ensemble approaches to enhance understanding, encourage further research, and promote practical implementation in both text and code generation. We categorize LLM ensembles into seven main methods—weight merging, knowledge fusion, mixture-of-experts, reward ensemble, output ensemble, routing, and cascading—analyzing capabilities of those approaches. Our findings highlight key benefits such as improved diversity representation, enhanced output quality, and greater application flexibility. These insights aid model selection for real-world tasks and crucially, lay groundwork for extending ensemble strategies to multimodal LLMs.
\end{abstract}

\begin{IEEEImpStatement}
Generative pretrained transformers (GPTs) are widely used large language models (LLMs) for text and code generation, providing researchers with an adaptive fine-tuning framework that alleviates the burden of model training. However, recent studies indicate that the best LLMs achieve only 57\% in chat generation quality due to their single architectural limitations and inability to capture diverse language patterns. This survey reviews LLM ensemble methods that address these shortcomings, resulting in a notable increase in instruction-following accuracy to 65\%. These ensemble techniques enhance diversity, output quality, and flexibility, making them suitable for various applications such as reasoning, question-answering, and code generation. The insights gained from this research are crucial for selecting effective models for real-world tasks. Additionally, LLM ensemble models can transform decision-making processes across sectors like medicine, finance, education, and customer service, enabling businesses to optimize operations, improve user experiences, and reduce costs associated with traditional models.
\end{IEEEImpStatement}


\section{Introduction}
\label{sec:intro}
\IEEEPARstart{M}{any} language processing applications are framed as generative modeling tasks, such as instruction following, which predicts conversational word sequences. This article focuses on text and code generation tasks, which encompass diverse research areas like mathematical reasoning, question answering (QA), massive multitask language understanding (MMLU), and code generation. These fields share the goal of generating language features from varied inputs.

Models based on the Generative Pretrained Transformer (GPT) architecture are widely used for text and code generation \cite{chen2021codex,instructgpt_ouyang2022}. (In this paper, "GPT" refers to this architectural concept unless a specific model like GPT-4o is named.) These GPT-based LLMs utilize embeddings to represent language features and can generate highly intelligible, natural text \cite{instructgpt_ouyang2022,Claude35_Anthropic2024,alpacaeval2023}. However, their responses often exhibit noticeable biases \cite{CalibratingLM_Jiang2023} and inconsistencies, a sensitivity largely attributed to biased training data \cite{switchtransformer_fedus2022}. This limitation in individual models motivates the exploration of ensemble approaches.

In typical GPT-based instruction following, for example, models are fine-tuned on conversational texts, with attention mechanisms characterizing embedding transitions through layers \cite{instructgpt_ouyang2022}. Training optimizes parameters via criteria like cross-entropy and human feedback. During synthesis, a meta-generation algorithm predicts likely language features \cite{metagenalgo_welleck2024, learningsummarizehumanfeedback_stiennon2022}. While this process improves robustness, the inherent limitations of relying on a single LLM persist, underscoring the need for alternative strategies like ensembling.

Individual LLMs face two key limitations. Firstly, their fixed language parameters can cause output inconsistencies \cite{llmcascades_yue2024} and restrict diverse language expression due to inherent biases \cite{calibratebeforeuse_zhao2021}. We attribute this issue stems from two assumptions about efficient information retrieval within fixed context windows and implicit knowledge representation without external access. Consequently, generated outputs can be model-dependent, leading to hallucinations and degraded text quality \cite{lmarefewshotlearners_brown2020}. Various techniques address these issues, including improved model architectures (e.g., Mixture-of-Experts \cite{shazeer2017outrageouslylargeneuralnetworks}, model herds \cite{llama3_dubey2024}, multimodal LLMs \cite{GPT4_OpenAI2023,gemini_geminiteam2024}), enhanced training criteria like knowledge distillation \cite{knowledgedistilminillm_gu2024}, and modified meta-generation algorithms incorporating intermediate decoding steps \cite{intermediatesteps_merrill2024}. Secondly, the closed-source nature of many powerful LLMs hinders data integration and raises privacy concerns, limiting industrial applications.

Since 2022, LLM ensemble learning has rapidly gained prominence in machine learning (ML) research \cite{modelsoups_ave_weights,weight_ave_ilharco2022}. These techniques combine multiple LLMs to improve performance in tasks such as feature generation, pattern recognition, and classification. Researchers have explored methods at both the architecture-level—such as weight merging \cite{weight_ave_ilharco2022}, knowledge fusion \cite{knowledgefusionwan2024}, mixture-of-experts (MoE) \cite{mixtralexperts_jiang2024}, and reward ensembles \cite{rewardmodelensembles_coste2024}—and the model-level, including output ensembles \cite{mixtureofagents_wang2024}, routing \cite{routerbench_hu2024}, and cascading \cite{llmcascades_yue2024}.

While MoE is a specific transformer architecture with multiple expert layers—often outperforming larger single LLMs like LLaMA 2 70B with fewer parameters \cite{mixtralexperts_jiang2024}—it can also serve as a base model within broader LLM ensemble frameworks (e.g., output ensembles). LLM ensembles, though not strictly adhering to traditional methods like bagging or boosting, exhibit similar functionalities: weight merging, for instance, resembles bagging in reducing variance \cite{weight_ave_ilharco2022, bagging_breiman96}, while cascading can be seen as a form of boosting \cite{llmcascades_yue2024, adaboost_FREUND1997119}. Given these functional parallels, we categorize these approaches as "LLM ensemble" in this article.

\begin{figure}[t!]
\centering
\includegraphics[width=0.48\textwidth]{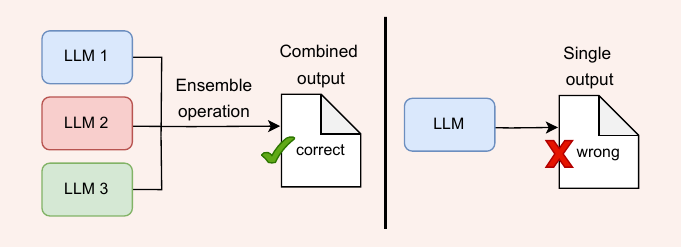}
\caption{The difference between an ensemble LLM and a single LLM. The single LLM can limit generalization ability. }
\label{fig:llmensemble_vs_singlellm}
\end{figure}

LLM ensembling has shown notable success, for instance, in code generation where multiple LLMs are used to evaluate code snippet predictions more effectively than a single LLM \cite{mixtralexperts_jiang2024, routerbench_hu2024, purifyingllms_li2024} (see Figure \ref{fig:llmensemble_vs_singlellm}). Recently, these techniques have also been extensively applied to text generation to address limitations of conventional approaches \cite{weight_ave_ilharco2022,modelsoups_ave_weights, mixtralexperts_jiang2024, ave_weights_ibm,taskarithmetic,adamerging,ties_merge_yadav2023,dare_yu2024,packllms_mavromatis2024,mergekit_goddard2024,checkpointmergingbayesoptim_liu2024, dispersethenmerge_fu2024,modelmergingsafetyalignment_hammoud2024, DPPApruning_zhu2024,prometheus2_kim2024,mitigatingalignmenttaxrlhf_lin2024,concretesubspacelearningbased_tang2023,llmensembleforEcommerce_fang2024,GAMoEs_akiba2024,knowledgefusionwan2024,fusechat_wan2024,ontheflyfusionllms_hoang2024, tokenprobfusion_yu2024, rewardmodelensembles_coste2024,warm_ramé2024,lowrankadoptation_zhai2023,multi-headRM_ahmed2024,helpingherdingrewardmodel_eisenstein2024,rewardmodelensembleRlhf_zhang2024, mixtureofagents_wang2024,fusingmodels_wang2024,llmblender_jiang2023,urg_lv2024,gettingmore_si2023,votellms_oniani2024, llmrouting_shnitzer2024,rewardrouter_lu2023,hybridllm_ding2024,harnessingpowermultipleminds_srivatsa2024,routoo_mohammadshahi2024,routerbench_hu2024, llmcascades_yue2024,frugalgpt_chen2023,languagemodelcascades_dohan2022}. While differing from code generation applications in data and evaluation specifics, these text generation ensembles utilize diverse model structures. Some focus on enhancing mechanisms such as weight merging \cite{weight_ave_ilharco2022,modelsoups_ave_weights, ave_weights_ibm, taskarithmetic, adamerging, ties_merge_yadav2023,dare_yu2024,packllms_mavromatis2024,mergekit_goddard2024}, knowledge fusion \cite{knowledgefusionwan2024,fusechat_wan2024}, or reward modeling \cite{rewardmodelensembles_coste2024,warm_ramé2024,lowrankadoptation_zhai2023,multi-headRM_ahmed2024} with source LLMs, or improving MoE gating functions \cite{switchtransformer_fedus2022,MoEsinstructiontuning_shen2023}. Others directly combine LLM outputs \cite{mixtureofagents_wang2024,fusingmodels_wang2024,llmblender_jiang2023,urg_lv2024,gettingmore_si2023,votellms_oniani2024} or model the end-to-end routing process from prompt to LLMs \cite{routerbench_hu2024,hybridllm_ding2024,harnessingpowermultipleminds_srivatsa2024,Fly-SwatorCannon_akota_2024,rewardrouter_lu2023,routoo_mohammadshahi2024,llmrouting_shnitzer2024,llmcascades_yue2024,frugalgpt_chen2023}.

LLM ensemble techniques generally outperform single GPT-based models by better representing diversity among LLM agents and their output features. These methods effectively combine varied strengths to enhance responses, particularly in tasks like question answering, utilizing both architecture-level (e.g., MoEs) and model-level (e.g., output ensemble, routing) frameworks. The practice of human developers using hierarchical structures to optimize code from multiple solutions \cite{cracking_mcdowell2015} further underscores the relevance and need for a comprehensive review of LLM ensemble models, especially for code generation applications.

This survey defines and describes LLM ensemble techniques using original categorizations and subjectively evaluates selected models for generation performance. For each technique, we focus on two to three papers that demonstrate representative performance and broader application potential, analyzing them for text and code generation across metrics, model structures/training, and cross-task comparisons. Our review aims to enhance understanding of current techniques, encourage further research and practical implementation, and ultimately broaden the real-world application of LLM ensembles, overcoming single-LLM limitations.

While existing surveys cover related areas—such as Lu et al. \cite{mergeensemblecooperatesurvey_lu2024} on LLM collaboration and Yang et al. \cite{modelmergingllms_yang2024} on broader ensemble techniques across machine learning fields including multimodal LLMs—our work offers a more concentrated focus. This survey specifically examines LLM ensemble methods for text and code generation, providing in-depth insights relevant to practical applications in these domains.

This article first reviews the conventional transformer framework for language generation, including GPT-based discriminative and generative tasks. We then analyze the limitations of these single-model approaches and introduce key LLM ensemble techniques, such as MoE, output ensembles, and routing. Finally, we survey emerging language generation methods using LLM ensembles, discussing their motivations, implementations, and highlighting remaining challenges and future research directions.

\section{Conventional Text Generation Using an LLM}
\label{sec:background}

The transformer model, introduced in 2017 \cite{transformer_vaswani2017}, marked a significant advancement in natural language generation (NLG). 
This model is versatile, supporting various generative tasks across time series analysis, vision, and image processing. 
In NLG, transformers model the relationship between text and its linguistic realizations, enhancing performance across a broad range of applications.
For instance, BERT utilizes bidirectional encoder representations to better understand sentence context \cite{bert_kenton2019}, Transformer XL extends the ability to learn dependencies beyond fixed lengths \cite{transformerxl_dai-etal-2019}, XLNet improves upon these approaches with permutation-based training that maintains bidirectionality while avoiding prediction-perception discrepancy \cite{xlnet_yang2019}, Longformer addresses document-level tasks by implementing an attention mechanism that scales linearly with sequence length \cite{longformer_beltagy2020}, and T5 treats all NLP tasks as text-to-text problems, facilitating its application in translation, summarization, and question answering \cite{t5_chung2022}.

\begin{figure}[htb!]
  \centering
  \includegraphics[width=0.48\textwidth]{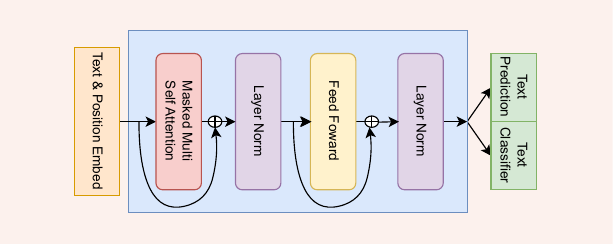}
  \caption{Architecture of a Transformer as a backbone of LLM, which is fine-tuned on a specific task given the text and position embeddings.}
  \label{fig:LLM-architecture}
\end{figure}

Among these, the context-dependent transformers, specifically fine-tuned for particular tasks, have shown the success in NLG \cite{gpt_radford2018}. 
The Generative Pretrained Transformer (GPT) is a prominent example architecture of such models, known as a large language model.
GPT-based models in language tasks can generate or classify a sequence of text using attention mechanism layers followed by blocks with high-dimensional inner states.
The attention mechanism computes a weighted representation of input sequences, focusing on relevant contexts and dependencies to enhance language understanding and generation.
Each block contains a feed-forward neural network (FFN) that processes the output from the attention mechanism to learn complex representations.
An architectural example of a Transformer as a backbone of LLM is illustrated in Figure \ref{fig:LLM-architecture}.
The GPT architecture involves text and position embeddings that represent observed features, which are inputted into the transformer, and the transitions between layers are characterized by attention weights.

GPT-based language generation produces highly intelligible and natural texts.
Compared to traditional recurrent-network-based methods, this attention-based approach is more efficient at handling long-range dependencies in language features, which correlate well to human perception. 

Additionally, input transformation and fine-tuning techniques are used to train these models on various tasks, enhancing the diversity of their transfer performance
\cite{sparsetransformer_child2019,chen2021codex,lmarefewshotlearners_brown2020,lmareunsupervisedmultitasklearners_radford2019,GPT4_OpenAI2023}. 
The model pre-training is typically done using a next token prediction approach.
At the training stage, a high-capacity language model with parameters $\Theta$ is first trained on a large corpus of text tokens $\mathcal{U}= \{  u_1,...u_n \}$ by maximizing the likelihood as

\begin{equation}
L_1(\mathcal{U}) = \sum_i \text{log}\ P(u_i\ |\ u_{i-k},..., u_{i-1}; \Theta )
\label{eq:lmobjective}
\end{equation}
where $P$ represents the probability of the current token given the previous tokens \cite{gpt_radford2018}.
Given a sequence of input tokens, $x^1,...,x^m$, the final transformer block’s activation $h$ is obtained. 
Next, a fine-tuning stage follows and adopts the pretrained model to a discriminative or generative task by feeding $h$ and parameters $W_y$ into an added linear output layer to predict the label $y$

\begin{equation}
P(y|x^1,...,x^m) = \texttt{softmax}(h\ W_y).
\end{equation}

Then, the following objective is often maximized for supervised fine-tuning as

\begin{equation}
L_2(\mathcal{C}) = \sum_{x,y} \text{log}\ P(y|x^1,...,x^m) 
\end{equation}
where $\mathcal{C}$ is a labeled dataset.
The fine-tuning objective that includes the language modeling objective typically improves generalization of the supervised model and accelerate convergence \cite{semisupervisedmultitasklearningsequence_rei2017,semisupervisedsequencetaggingbidirectional_peters2017}.
Therefore, one of the most efficient fine-tuning objective can be considered a linear combination of the language modeling objective (\ref{eq:lmobjective}) as 

\begin{equation}
L_3(\mathcal{C}) =  L_2(\mathcal{C}) + \lambda * L_1(\mathcal{C})
\end{equation}
where $\lambda$ is the weight parameter \cite{gpt_radford2018}.


The GPT-based language understanding system is typically designed to tackle either discriminative or generative tasks, and its capability to generate human-like responses is limited.
To enhance conversational capabilities, multiple approaches are often integrated.
Instruction tuning - fine-tuning models on diverse instruction-following tasks - has emerged as a foundational technique \cite{instruction-tuning_wei2022}.
Building on this foundation, reinforcement learning from human feedback (RLHF) provides further refinement \cite{instructgpt_ouyang2022}. 
This multi-stage process begins with demonstration data where human experts provide exemplary responses to prompts. 
Next, a reward model is trained on human-ranked outputs to evaluate response quality.
Subsequently, a proximal policy optimization algorithm \cite{proximalpolicyoptimizationalgorithms_schulman2017} is used to refine the policy based on the rewards determined by the reward model.

For language generation, the process begins with a sequence of input text tokens, which includes specific \textit{start} and \textit{end} tokens. These tokens are crucial for generating the embedding features needed for the language generation process.

The next section reviews several recent approaches based on LLM ensemble techniques for overcoming the limitations of the single-LLM-based approach described in Section \ref{sec:intro} and improving language generation. 
The review includes some mathematical details that are uncommon in the general ML literature but essential for using these models in language generation.

\section{Primary Methods for LLM Ensembling}
\label{sec:basicllmensembling}

\begin{figure}[htb!]
  \centering
  \includegraphics[width=0.48\textwidth]{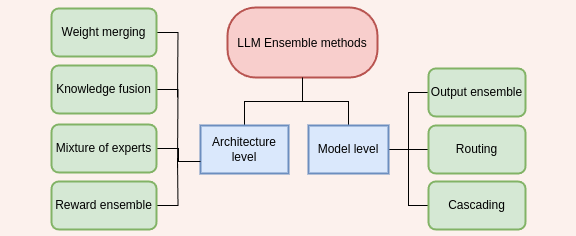}
  \caption{LLM ensemble methods in our categorization. We identify seven primary approaches at the architecture or model level.}
  \label{fig:llmensemblediagram}
\end{figure}

Since 2022, LLM ensemble learning techniques have been effectively applied to model language features across diverse applications, such as extracting e-commerce product features \cite{llmensembleforEcommerce_fang2024}, generating image captions \cite{LDRE_yang2024}, and classifying medical diagnoses \cite{diagnostic_Barabucci_2024}. A key benefit of ensembling is its capacity to generalize from diverse outputs by combining knowledge through various frameworks, either at the architecture level (e.g., weight merging, knowledge fusion) or the model level (e.g., output ensemble, routing), as illustrated in Figure \ref{fig:llmensemblediagram}. Fundamentally, LLM language generation is a next-token prediction task, and ensemble language modeling aims to enhance the quality of these predicted tokens. This section reviews foundational ensemble techniques from this perspective.

\subsection{Architecture-Level Ensemble}
\label{sec:arch-level}

\subsubsection{Weight Merging}

Weight merging combines the parameters (weights) of multiple trained LLMs into a single model. This architecture-level technique operates directly on the model parameters, typically requiring the source LLMs to share compatible architectures, and aims to consolidate their diverse learned knowledge.

\begin{figure}[htb!]
  \centering
  \includegraphics[width=0.48\textwidth]{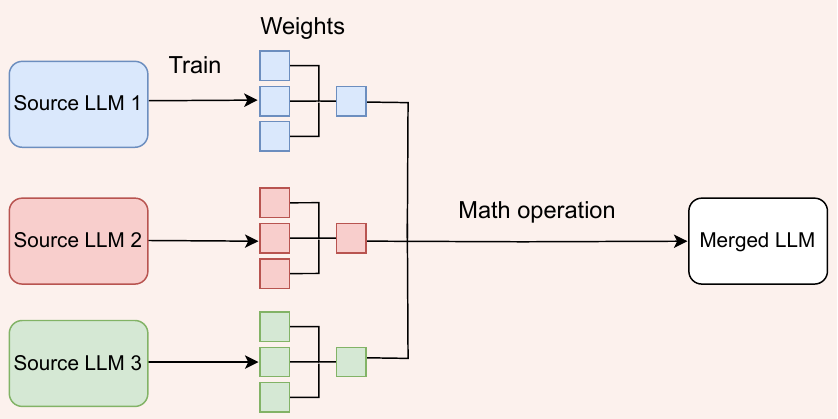}
  \caption{Weight merging with three source code of LLMs. Weights are typically operated after training models.}
  \label{fig:weightmerging}
\end{figure}

Traditionally, weights $\text{W}i$ from $K$ LLMs are merged, often by averaging: 
\begin{equation}
\text{W} = \frac{1}{K} \sum_{i=1}^K \text{W}_i
\label{eq:weightave}
\end{equation}
where $\text{W}$ represents the final model's weights (Figure \ref{fig:weightmerging} illustrates this with three source LLMs). $K$ can denote all LLMs \cite{ave_weights_ibm} or a subset, such as top-$K$ models selected via a greedy algorithm \cite{modelsoups_ave_weights}. Furthermore, $\text{W}$ can be interpolated with pre-trained weights using a parameter $\alpha$ \cite{weight_ave_ilharco2022}. These weight averaging methods, by interpolating between models, have been shown to improve accuracy on downstream tasks.

\begin{figure}[!htb]
  \centering
  \includegraphics[width=0.48\textwidth]{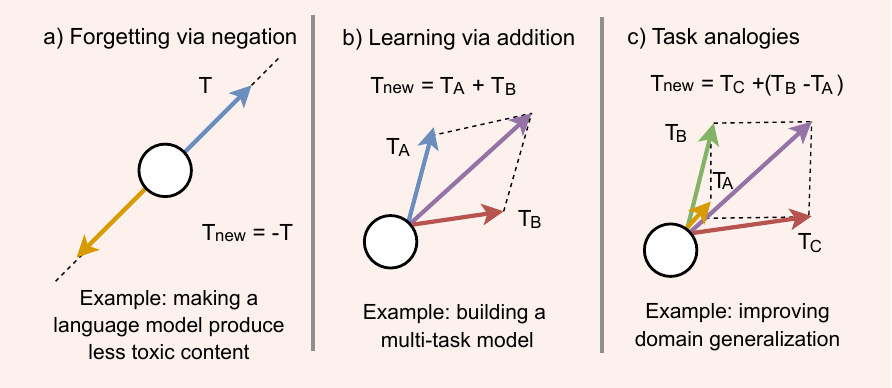}
  \caption{Arithmetic operations with a task vector $T$. $T$ is edited to improve model performance \cite{taskarithmetic}.}
  \label{fig:tascvector}
\end{figure}

Building on these findings, extended merging methods extrapolate weights by utilizing task vectors \cite{taskarithmetic,adamerging}. A task vector $T$ is typically defined as the difference between fine-tuned weights ($\text{W}_{ft}$) and pre-trained weights ($\text{W}_{pre}$) of an LLM: 
\begin{equation} 
T = \text{W}_{ft} - \text{W}_{pre} 
\end{equation} This task vector $T$ can then be transformed (denoted $T_{new}$) using various techniques such as arithmetic operations \cite{taskarithmetic}, trimming and sign-based selection (TIES) \cite{ties_merge_yadav2023}, or random dropping and re-scaling (DARE) \cite{dare_yu2024}, as illustrated for arithmetic operations in Figure \ref{fig:tascvector}. The final merged weights $\text{W}$ are obtained by applying this transformed task vector to a base pre-trained LLM's weights $\text{W}_{pre}$ with a scaling factor $\lambda$, often determined using a validation set: 
\begin{equation} 
\text{W} = \text{W}_{pre} + \lambda \ T_{new} 
\end{equation}

Another approach, perplexity-based merging, performs test-time aggregation by combining output logits using importance weights \cite{packllms_mavromatis2024}. In this method, each LLM's expertise is first measured by its perplexity on an input prompt, and a greedy algorithm then minimizes perplexity to derive these importance weights. This extends weight merging to inference time, creating a training-free ensemble that can easily incorporate newly released open-source LLMs. Recognizing the expanding research in model merging, MergeKit offers a comprehensive open-source library to support these strategies \cite{mergekit_goddard2024}.

\subsubsection{Knowledge Fusion}
\begin{figure}[htb!]
  \centering
  \includegraphics[width=0.48\textwidth]{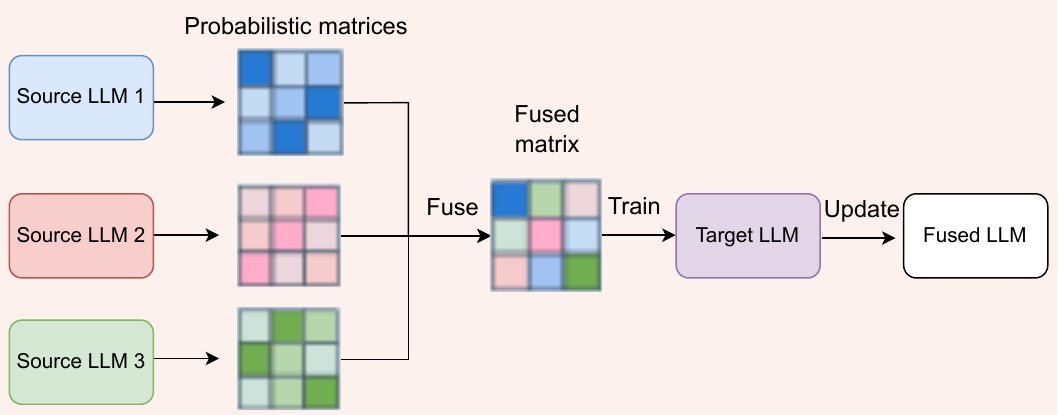}
  \caption{Fusing three LLMs. Fused knowledge is used to train a target LLM.}
  \label{fig:knowledgefusion}
\end{figure}

Knowledge fusion, an architecture-level technique, aims to consolidate the collective knowledge of multiple source LLMs into a target LLM by defining fusing functions over their probabilistic output distributions \cite{knowledgefusionwan2024}. Figure \ref{fig:knowledgefusion} illustrates this process using three source LLMs to train a target LLM. This method extends traditional knowledge distillation \cite{knowledgedistilling_hinton2015}, which transfers knowledge from a larger teacher model to a smaller student model, primarily for tasks like text classification, to create more efficient student models.

In knowledge fusion, the objective is to enhance a target LLM by training it with a fused probabilistic distribution matrix $\textbf{P}_q$. For each text sample $q$ from a corpus $C$, $\textbf{P}_q$ is derived by combining the token-level prediction distributions $\textbf{P}^{\theta_j}_q$ from $K$ source LLMs (where $\theta_j$ denotes the parameters of the $j$-th LLM): 
\begin{equation}
\textbf{P}_q = \text{Fusion(} \textbf{P}^{\theta_1}_q,\ \textbf{P}^{\theta_2}_q ,...,\ \textbf{P}^{\theta_K}_q \text{)}
\end{equation}
The $\text{Fusion}(\cdot)$ function can employ various strategies, such as selecting the output based on the minimum cross-entropy or a weighted average of cross-entropy scores from the source models \cite{knowledgefusionwan2024}. During the training of the target LLM, the capabilities of the source LLMs are transferred using the following objective function: 
\begin{equation}
\begin{split}
L &= \lambda\ L_{Target} + (1 - \lambda)\ L_{Fusion} \\
L_{Target} &= - \mathbb{E}_{q \sim C}\ [\mathbb{D}(\textbf{Q}_q,\ \textbf{O}_q)] \\
L_{Fusion} & = - \mathbb{E}_{q \sim C}\ [\mathbb{D}(\textbf{Q}_q,\ \textbf{P}_q)]
\end{split}
\label{eq:knowledge-fusion-loss}
\end{equation}
Here, $\mathbb{D}(\cdot)$ is a discrepancy function (e.g., KL divergence) between two matrices, $\textbf{Q}_q$ is the output distribution matrix of the target LLM, $\textbf{O}_q$ is the one-hot label matrix, and $\lambda$ is a weighting parameter.

This knowledge fusion method can be extended to a pairwise approach, particularly useful when source LLMs differ in size from the target LLM. For each source LLM, designated as a pivot, pairwise fused matrices are computed between this pivot and every other source LLM \cite{fusechat_wan2024}. The pairwise fused matrix for the $j$-th LLM (when the $v$-th LLM is the pivot), $\textbf{P}^j_q$, is derived as: 
\begin{equation}
\textbf{P}^j_q = \text{Fusion(} \textbf{P}^{\theta_v}_q,\ \textbf{P}^{\theta_j}_q \text{)}\ |_{v \neq j}.
\end{equation}
The pairwise fusion objective for the target LLM, incorporating this specific $\textbf{P}^j_q$, is then defined as: 
\begin{equation}
L_{Fusion} = - \mathbb{E}_{q \sim C}\ [\mathbb{D}(\textbf{Q}_q,\ \textbf{P}^j_q)]
\end{equation}
This $L_{Fusion}$ is subsequently substituted into the overall loss function $L$ in Equation (\ref{eq:knowledge-fusion-loss}) to compute the final training objective.

Different LLMs often employ distinct tokenizers, leading to variations in their vocabulary matrices. For example, one LLM's tokenizer might output "cryptocurrency" as a single token, while another's might produce "crypto" and "currency" as separate tokens. This discrepancy in vocabulary representations complicates the mathematical fusion process and is known as the token alignment problem \cite{token-align-slm_fu23d}.

To address this token alignment issue, dynamic programming can be employed to find a one-to-one exact mapping that minimizes the total cost of transforming one token sequence into another \cite{token-align-slm_fu23d}. Alternatively, a minimum edit distance approach aligns tokens from different tokenizers by minimizing the necessary edits, thereby preserving crucial information within the vocabulary distribution \cite{knowledgefusionwan2024}.

Alternatively, an embedding representation approach tackles token alignment by combining collaboration weights with relative representation matrices from LLMs. These matrices capture token similarities, often anchored by reference tokens, and merging them with collaborative model weights enhances generalization for tasks like word completion \cite{deepen_huang2024,tokenalignByEmb_xu2024}.

\subsubsection{Mixture of Experts}
\label{sec:MoE}


Architecture-level ensemble methods, known as Mixture of Experts (MoE), are employed in transformers to model multidomain data distributions for tasks such as text generation, mathematics, code generation, and natural language understanding (NLU) \cite{mixtralexperts_jiang2024, switchtransformer_fedus2022, MoEsinstructiontuning_shen2023}.

\begin{figure}[ht]
  \centering
  \includegraphics[width=0.48\textwidth]{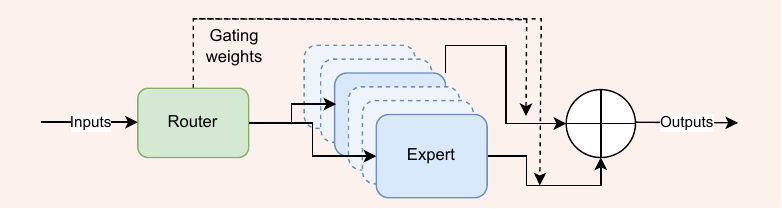}
  \caption{Mixture of experts layer. Each input token is typically routed to suitable experts to predict a next token.}
  \label{fig:mixture-of-experts-layer}
\end{figure}

For an input token $x$ and $K$ expert networks $E_i(x)$, the output $y$ of an MoE layer is the weighted sum of expert outputs, with weights determined by a gating network (router) $G(x)i$ (see Figure \ref{fig:mixture-of-experts-layer}). This is expressed as: 
\begin{equation}
\text{y} = \sum_{i=1}^{K} G(x)_i \cdot E_i(x)
\label{eq:MoEs}
\end{equation}

MoE combines expert distributions from diverse datasets via a gating mechanism, creating more optimal distributions than individual experts \cite{mixtralexperts_jiang2024}. This allows for effective model scaling without significantly increasing computational complexity \cite{gshardscalingMoE_lepikhin2020}.

In Equation \ref{eq:MoEs}, $G(x)_i$ represents the top $n$ experts ($n > 1$), ensuring meaningful gradients for the gating functions \cite{shazeer2017outrageouslylargeneuralnetworks}. $E_i(x)$ denotes the $n$ expert sub-blocks, potentially with distinct weights (e.g., SwiGLU architecture \cite{swiglu_shazeer2020}).

Increasing the number of experts per token ($n$) raises the model's parameter count with constant computational costs, due to sparse computation in MoE \cite{mixtralexperts_jiang2024}. Alternatively, Switch Routing selects a single expert ($n=1$) from a reduced batch of parallel experts, simplifying implementation and lowering routing and communication costs \cite{switchtransformer_fedus2022}.

\subsubsection{Reward Ensemble}
\label{sec:rewardensemble}

Architecture-level ensembling using multiple reward models (RMs), termed a reward ensemble, can be applied to reward modeling in reinforcement learning. This section discusses reward ensembles within reinforcement learning from human feedback (RLHF).

\begin{figure}[ht]
  \centering
  \includegraphics[width=0.48\textwidth]{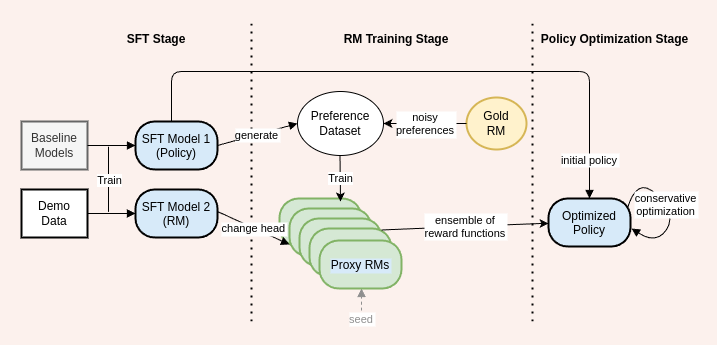}
  \caption{RLHF pipeline in reward ensemble. Multiple RMs are used to mitigate reward hacking.}
  \label{fig:RLHF_pipeline}
\end{figure}

This approach was introduced to address reward hacking (overoptimization), where a policy achieves high rewards while disregarding true human preferences \cite{rewardmodelensembles_coste2024, rewardhacking_amodei2016}. The typical structure of a reward ensemble in the RLHF pipeline is depicted in Figure \ref{fig:RLHF_pipeline}.

In this model, learned RMs approximate human preferences, represented by the policy $\pi$, which is optimized relative to the policy model and RMs. This optimization can involve LLMs with different initialization seeds \cite{rewardmodelensembles_coste2024} or low-rank adaptation during supervised fine-tuning (SFT) or pretraining \cite{lowrankadoptation_zhai2023, helpingherdingrewardmodel_eisenstein2024}.

The initial policy, denoted as $\pi_{init}$, and the optimized policy $\pi$ together create the gold RM score $R$. This score is calculated based on the Kullback–Leibler divergence $D_{KL}(\pi\ ||\ \pi_{init})$, which varies depending on the optimization method used.
If $R(0) := 0$ by convention and $\alpha$ and $\beta$ are parameters, a reward function of RM can be written as 
\begin{equation}
R(d) = d(\alpha - \beta \log d)
\end{equation}
where $d := \sqrt{D_{KL}(\pi\ ||\ \pi_{init})}$. 
Using the Mean Optimization method \cite{convexboyd2004}, the ensemble reward function $R$ for $K$ LLMs is: 
\begin{equation}
R(q, a) := \frac{1}{K} \sum_{i=1}^K R_i(q, a)
\label{eq:meanrewardfunction}
\end{equation}
where $R_i(q, a)$ is the gold RM score for prompt $q$ and its response $a$ sampled from the $i$-th RM. Besides Mean Optimization, other techniques can be evaluated based on performance criteria, such as Worst-Case Optimization \cite{convexboyd2004} and Uncertainty-Weighted Optimization \cite{Disagreement-RegularizedBrantley2020}.

Due to the computational cost of RM ensembling, a multi-head RM approach, which uses shared encoders with separate linear heads, has been proposed to save memory and time during LLM training \cite{multi-headRM_ahmed2024}.

The RM ensemble can be extended to include weight-averaged reward models (using Equation \ref{eq:weightave}) and by fine-tuning multiple RMs with varying hyperparameters. This enhances scalability, improves robustness against label corruption, and increases reliability during distribution shifts, while also mitigating reward hacking \cite{warm_ramé2024}.

\subsection{Model-Level Ensemble}
\label{sec:model-level-ensemble}
\subsubsection{Output Ensemble}
Output ensemble is a technique that does not require training LLMs. It combines responses from several LLMs to enhance performance, accuracy, and robustness compared to using a single LLM.

In a classification task, output ensemble can involve voting or classifying outputs from multiple LLM agents. Voting means choosing the most common output from the responses of different LLMs.
The final output label $\textbf{y}$ of majority voting over the LLMs' outputs can be written as
\begin{equation}
\textbf{y} \sim \# argmax\ \{ i\ |\ y_i = \textbf{y}\}
\end{equation}
where $y_i$ is the output label by $i$-th agent.
In classifying outputs, a fully connected neural network can be trained to ingest the responses of all LLMs to produce a class prediction \cite{fusingmodels_wang2024}.

For a text generation task, output ensemble can be applied to combine output texts by prompting an LLM to synthesize them. 
Given an input prompt $x$, the synthesized text $y$ over the $n$ LLM outputs can be written as 
\begin{equation}
y = \oplus_{j=1}^n [A_{j} (x)] + x
\label{eq:MoA}
\end{equation}
where $A_j(x)$ is the output text from $j$-th LLM, $\oplus_{j=1}^n [A_{j} (x)]$ represents application of the synthesize prompt for fusing $n$ output texts, and + means concatenation of texts \cite{mixtureofagents_wang2024}.
The formula \ref{eq:MoA} represents the process of generating and synthesizing responses from $n$ LLMs.

The output ensemble has been effectively used to rank and fuse the top-$K$ generated texts. This approach takes advantage of the strengths of each base LLM. Research shows that this post-ranking fusion framework can better capture the differences between agent responses \cite{llmblender_jiang2023}. It also produces better results by utilizing the top-$K$ outputs \cite{llmblender_jiang2023, urg_lv2024}.

In this framework, various transformer-based ranking models can be trained and tested using the CE criterion. Examples include PAIRRANKER \cite{llmblender_jiang2023}, URG \cite{urg_lv2024}, and SummaReranker \cite{summareranker_ravaut2022}. These ranking models typically use a scoring system to evaluate and rank the outputs, leading to a sorted list of results.

\textbf{Scoring model. }
A scoring model produces meaningful numeric values by either classifying data with an ML model or generating results with a language model. These models are commonly used in ranking and various text generation tasks, such as outputs ranking, instruction-following \cite{llmblender_jiang2023}, QA reasoning \cite{gettingmore_si2023}, and image caption generation \cite{LDRE_yang2024}.

For instance, a random forest (RF) classifier can be trained to give confidence scores for answers generated by LLMs. This helps in choosing the most confident response \cite{gettingmore_si2023}. 
Additionally, scoring models are important for routing tasks.





\subsubsection{Routing} 

\begin{figure}[ht]
  \centering
  \includegraphics[width=0.48\textwidth]{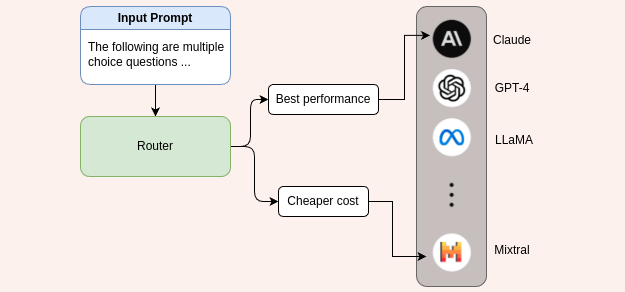}
  \caption{Routing system with LLMs of different cost and performance. It routes the input query to a corresponding LLM depending on preferences.}
  \label{fig:router}
\end{figure}

Routing is a technique that directs input queries to multiple LLMs. The design of a multi-LLM router is illustrated in Figure \ref{fig:router}.
In this system, the router assigns a score to each input query using a scoring model. It then sends the query to the most appropriate LLM based on that score. The scoring function varies depending on the design and the specific tasks.

For general tasks like QA, summarization, and information extraction, a BART score \cite{bartscore_yuan2021} can evaluate the quality of responses from different LLMs. This score is effective because it correlates well with the actual answers \cite{hybridllm_ding2024}.

For tasks involving mathematical or natural language reasoning, a confidence score from a transformer-based classifier, such as RoBERTa or T5, can help identify the most reliable LLM to use \cite{harnessingpowermultipleminds_srivatsa2024}.

This classifier-based routing can be extended to different fields by training multiple classifiers. For instance, in finance, legal, and general knowledge tasks, the output from a kNN classifier $g_m(\cdot)$ can serve as a correctness score for the LLM:
\begin{equation}
g_m(x_i^{d}) = \frac{1}{k} \sum_{e\ \in\ \text{NN} (\ \phi(x_i^{d}),\ k,\ D)} y\ (e, m)
\label{eq:knnrouter}
\end{equation}
where $x_i^{d}$ is a sample from a domain dataset $d$, $y\ (e, m) \in \{ 0, 1 \} $ is the correctness of model $m$ on the embedded input $e$, and $\text{NN} (\ \phi(x_i^{d}),\ k,\ D)$ is the set of $k$ closest embedded neighbors from $D$ datasets to the new embedded sample $\phi(x_i^{d})$ \cite{llmrouting_shnitzer2024}.
$x_i^{d}$ can be embed using an embedding model, e.g., sentence transformer $\phi$ \cite{sentencebert_reimers2019}.

This classifier-based router can also incorporate the cost-performance trade-off \cite{routerbench_hu2024, Fly-SwatorCannon_akota_2024}.
For example, the kNN router in (\ref{eq:knnrouter}) can model a performance score using the willingness-to-pay parameter $\lambda$
\begin{equation}
\text{performance score}_{i,j} = \lambda \cdot P_{i,j} - \text{cost}_j
\end{equation}
where the kNN classifier estimate the predicted performance $P_{ij}$ of $j$-th LLM on sample $x_i$, cost is the total cost approximated using the cost per token metric, and higher $\lambda$ indicates a preference for superior performance at a higher cost \cite{routerbench_hu2024}. 

This routing technique can also improve the reward ensemble method for RLHF in Section \ref{sec:rewardensemble} by sampling from a routed LLM. This allows for training a single RM instead of multiple RMs \cite{rewardrouter_lu2023}.

Routers can be cost-effective models that simplify the process of connecting inputs to outputs. However, they struggle to save costs when handling complex reasoning tasks. For example, in mathematics, routers determine the difficulty of questions and the correctness of answers based only on their text descriptions. This reliance can lead to lower accuracy \cite{llmcascades_yue2024}. Additionally, fine-tuning routers using smaller language models may require them to learn from a large number of training samples, which can complicate the learning process.

\subsubsection{Cascading}

\begin{figure}[!htb]
  \centering
  \includegraphics[width=0.48\textwidth]{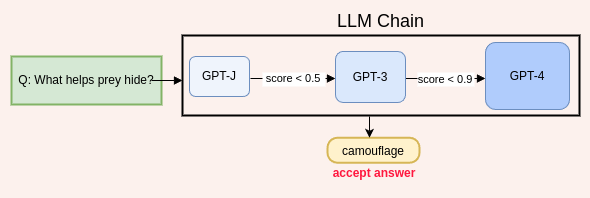}
  \caption{The cascade chain of three LLMs. Getting acceptable answers from weaker but cheaper LLMs can save the cost of invoking stronger costly LLMs.}
  \label{fig:cascade}
\end{figure}

To address the routing problems mentioned above, cascading techniques have been proposed. In this approach, a list of LLM APIs is called one after the other. A decision maker then chooses whether to use a response from a weaker LLM or a stronger one \cite{frugalgpt_chen2023}.
Figure \ref{fig:cascade} illustrates the typical cascade process, which involves chaining three LLMs. This includes a weaker, more affordable LLM and stronger, more expensive ones.

In this cascade, the weaker LLM is used first to generate an initial answer along with some metadata. This answer is then given to a decision maker. The decision maker evaluates whether the answer can be accepted as final, based on a score and a predetermined threshold. For instance, in mathematical reasoning tasks, an agreement score $s$ can measure the consistency of the $j$-th LLM:
\begin{equation}
    s = \frac{\sum_{i=1}^K \mathds{1}_{A_i^j = A^j}}{K}
\end{equation}
where $A^j$ is the answer from $j$-th LLM. 
If $s$ is higher than a threshold, $A^j$ is selected as the most consistent final answer among $K$ samples \cite{llmcascades_yue2024}.
If the system rejects the initial answer, it uses a more powerful LLM to find a better response. At the same time, it calculates the total cost of answering the question. This step-by-step approach can deliver performance similar to that of a stronger LLM compared to the one-step process in routing. It also helps to significantly lower costs.

\subsection{Prompt Engineering}
\label{sec:promptengineering}
Prompt engineering is a technique that improves the quality of output from a single LLM. Key methods include prompt ensemble and prompt augmentation. Prompt engineering is not an LLM ensemble technique, but we discuss some relevant approaches, as they can enhance the LLM ensemble results.

Prompt ensemble combines results from multiple prompts. In text generation tasks like multiple-choice (MC) questions, the sensitivity of LLMs to prompts \cite{promptsensitivity_lu2022} can be leveraged to create a diverse set of prompts. This diversity helps improve performance by paraphrasing templates and permuting examples. For a MC question with $n$ prompts, the predictive distributions from these prompts are averaged to determine the ensemble accuracy \cite{CalibratingLM_Jiang2023}.

\begin{figure}[!htb]
  \centering
  \includegraphics[width=0.48\textwidth]{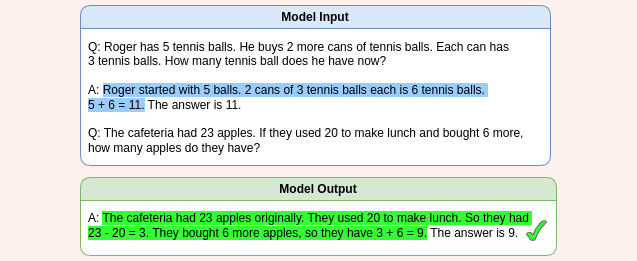}
  \caption{The input prompt of CoT with one intermediate step exemplar \cite{chainofthought_wei2023}.}
  \label{fig:chainofthought}
\end{figure}

Prompt augmentation involves adding relevant information to a prompt to boost reasoning ability. A well-known method is chain-of-thought prompting (CoT). CoT breaks down complex problems into smaller steps, providing examples for each step before arriving at the final answer \cite{chainofthought_wei2023}. An example of a CoT prompt with one intermediate step is illustrated in Figure \ref{fig:chainofthought}. Other methods, like program-of-thought prompting (PoT) \cite{programthoughts_chen2023} and retrieval-augmented prompting (RA) \cite{retrivalaugmentedprompts_si2023}, also enhance output quality.

LLMs with smaller parameter sizes can experience various performance issues, such as problems with semantic understanding, missing steps, and errors in symbol mapping. Therefore, LLMs should have a large number of parameters, ideally over 540 billion. Performance then improves when the LLM learns from a few input-output examples.

To help LLMs tackle problems harder than examples, an extension of CoT called least-to-most prompting (LtM) has been developed \cite{leasttomost_zhou2023}. This method solves sub-problems sequentially and augments the answers to the prompt in a divide-and-conquer approach. Compared to CoT, LtM often allows LLMs to generalize better on complex reasoning tasks.

\section{Text and Code Generation using LLM Ensemble Methods}
Given the success of LLM ensembles in various language understanding tasks, we believe this approach can also benefit real-world text and code generation. Our goal is to expand the application of LLM ensemble models in these areas, addressing the limitations of single-LLM approaches.

LLM ensemble learning is an emerging research area. Its application to text and code generation has only recently been explored, with several publications covering related topics such as instruction following \cite{llmblender_jiang2023, mixtureofagents_wang2024, dare_yu2024}, math reasoning \cite{taskarithmetic, GAMoEs_akiba2024,knowledgefusionwan2024,fusechat_wan2024,llmcascades_yue2024}, MMLU \cite{mixtralexperts_jiang2024,harnessingpowermultipleminds_srivatsa2024,routoo_mohammadshahi2024}, conversation \cite{rewardmodelensembleRlhf_zhang2024, lowrankadoptation_zhai2023}, and code generation \cite{ties_merge_yadav2023,dare_yu2024,knowledgefusionwan2024,fusechat_wan2024,olmoe_muennighoff2025,mixtureofagents_wang2024,mixtralexperts_jiang2024}.

We selected articles based on: (1) significance within each main ensemble category (Section \ref{sec:basicllmensembling}), (2) provision of clear metrics and quantifiable results, and (3) representation of recent (2023-2024) and seminal works. This ensures coverage of established techniques and promising emerging methods. Our analysis indicates that LLM ensembling consistently enhances the naturalness, accuracy, and quality of generated text and code. Following our categorization framework, we examine 2-3 representative papers from each main ensemble approach, discussing their performance, benchmark results, computational requirements, and strengths/weaknesses. This focused selection allows for an in-depth analysis of impactful ensemble strategies and a comprehensive overview of the field's evolution.

\subsection{Evaluation Metrics and Datasets}
To assess these ensemble methods, we first detail the standard metrics and datasets for text and code generation.

For text generation, common metrics include:
\begin{itemize}
    \item \textbf{Accuracy}: Correctness of responses.
    \item \textbf{Win Rate}: Comparative preference over other models.
    \item \textbf{BLEU}: N-gram overlap with references, e.g., for fluency.
    \item \textbf{Alignment Score}: Adherence to instructions/user intent.
    \item \textbf{Human Evaluation Score}: Subjective assessment of coherence, relevance, and quality.
\end{itemize}
Key datasets include MMLU (multitask understanding) \cite{mmlu_hendrycks2021}, GSM8K (math reasoning) \cite{GSM8K_cobbe2021}, AlpacaEval (instruction following) \cite{alpacaeval2023}, and MT-Bench (multi-turn conversation) \cite{mt-bench_zheng2023}.

For coding, metrics focus on correctness and efficiency:
\begin{itemize}
    \item \textbf{Pass@k}: Probability of a correct solution in $k$ attempts.
    \item \textbf{CodeBLEU}: Syntactic and semantic similarity for code.
    \item \textbf{Execution Time \& CPU Usage}: Code efficiency.
\end{itemize}
Prominent datasets are HumanEval \cite{chen2021codex} and MBPP (Mostly Basic Python Problems) \cite{mbpp_austin2021} for testing functional correctness.

\subsection{Performance Analysis by Method}
Using this evaluation framework, we now analyze the performance, resources, and trade-offs of seven key LLM ensemble methods.

\subsubsection{Weight Merging}
Recent weight merging methods efficiently combine specialized language models without requiring additional training or extensive resources. We examine results from two papers: \cite{dare_yu2024}, which evaluates primary approaches including Task Arithmetic \cite{taskarithmetic}, TIES-Merging \cite{ties_merge_yadav2023}, and DARE \cite{dare_yu2024} (Table \ref{tab:dare}), and \cite{GAMoEs_akiba2024}, which advances the field using genetic algorithms to achieve state-of-the-art results (Table \ref{tab:eo}).

\textbf{Key Findings.} Weight merging approaches demonstrate remarkable effectiveness across diverse tasks. DARE \cite{dare_yu2024} eliminates up to 99\% of parameter differences without performance degradation, while different techniques show complementary strengths: Task Arithmetic \cite{taskarithmetic} excels in math (66.34\% on GSM8K), TIES \cite{ties_merge_yadav2023} in code generation (37.80\% on HumanEval), and TIES+DARE balances both capabilities. Though all merged approaches still trail GPT-4's 85.3\% on conversation tasks. Meanwhile, Evolutionary Optimization \cite{GAMoEs_akiba2024} uses genetic algorithms to discover optimal combinations by evolving architectures and merging weights, enabling three weak 7B models to collectively outperform GPT-3.5 on Japanese math reasoning (52.0\% vs. 50.4\%).

\textbf{Resource Requirements.} Both methods feature minimal computational demands compared to fine-tuning. DARE \cite{dare_yu2024} operates on CPUs alone, while Task Arithmetic \cite{taskarithmetic} and TIES Merging \cite{ties_merge_yadav2023} require no training, functioning purely through weight manipulation. Evolutionary Optimization \cite{GAMoEs_akiba2024} needs only a single GPU for optimization, not training. Neither approach increases parameter count or requires additional training data, making them significantly more accessible than conventional model development techniques.

\textbf{Trade-offs.} Weight merging methods offer remarkable computational efficiency, enabling researchers with limited resources to combine specialized capabilities without training. DARE excels in parameter efficiency through sparsification, while Evolutionary Optimization automates merging discovery, outperforming GPT-3.5 with smaller models. However, these approaches are limited by source model capabilities and show task-dependent variations in performance. Despite these constraints, weight merging democratizes advanced model development while reducing computational costs.

\begin{table}[!htb]
\caption{Performance comparison of primary weight merging approaches on text and code generation tasks \cite{dare_yu2024}}
\tablefont%
\setlength{\tabcolsep}{2pt}
\begin{tabular*}{21pc}{@{} p{75pt} p{35pt}<{\raggedright\hangindent4pt} p{45pt}<{\raggedright\hangindent4pt} p{45pt}<{\raggedright\hangindent4pt} p{45pt}<{\raggedright\hangindent4pt} @{}}
\hline
Model & Type & HumanEval & AlpacaEval & GSM8K \\
\hline\\[-17pt]
&&&&\\
WizardLM-13B & 1 & 36.59\% & 67.20\% & 2.20\% \\
WizardMath-13B & 2 & -- & -- & 64.22\% \\
Task Arithmetic & 1 + 2 & 28.66\% & 67.04\% & \textbf{66.34\%} \\
TIES Merging & 1 + 2 & \textbf{37.80\%} & 68.63\% & 15.77\% \\
TIES + DARE & 1 + 2 & 36.59\% & 68.70\% & 36.16\% \\
GPT-4 (03/14) & -- & -- & \textbf{85.3\%} & -- \\
\hline
\multicolumn{5}{l}{}\\[-5pt]
\multicolumn{5}{@{}p{21pc}@{}}{\hspace*{9pt}Task Arithmetic \cite{taskarithmetic} excels in math, TIES \cite{ties_merge_yadav2023} in code, and TIES+DARE \cite{dare_yu2024} balances both, though all trail GPT-4 on AlpacaEval.}\\
\end{tabular*}
\label{tab:dare}
\end{table}

\begin{table}[!htb]
\caption{Performance comparison of Evolutionary-Optimization-based weight merging against source models and GPT-3.5 \cite{GAMoEs_akiba2024}}
\tablefont%
\setlength{\tabcolsep}{2pt}
\begin{tabular*}{21pc}{@{} p{75pt} p{40pt}<{\raggedright\hangindent4pt} p{60pt}<{\raggedright\hangindent4pt} 
p{60pt}<{\raggedright\hangindent4pt} @{}}
\hline
Model & Type & JP-LMEH  & MGSM-JA  \\
\hline\\[-17pt]
&&&\\
WizardMath-7B & 3 & 60.1  & 18.4\% \\
Shisa Gamma 7B & 4 & 66.1  & 9.6\% \\
Abel 7B 002 & 5 & 56.5  & 30.0\% \\
EO & 3 + 4 + 5 & \textbf{70.5}  & \textbf{52.0\%} \\
GPT-3.5 & -- & --  &  50.4\% \\
\hline
\multicolumn{4}{l}{}\\[-5pt]
\multicolumn{4}{@{}p{21pc}@{}}{\hspace*{9pt}EO with weak open-source models outperformed GPT-3.5, a commercial model, on Japanese datasets, showing a strong generalization ability.}\\
\end{tabular*}
\label{tab:eo}
\end{table}

\subsubsection{Knowledge Fusion}
Knowledge fusion combines complementary knowledge from language models with diverse architectures through lightweight continual training. Unlike weight merging (requiring architectural similarity) or aggregation methods (demanding parallel deployment), this approach supports fusion across structurally different models. Introduced in FuseLLM \cite{knowledgefusionwan2024} and extended in FusionChat \cite{fusechat_wan2024}, it shows promising results across benchmarks (Table \ref{tab:fusechat}).

\textbf{Key Findings.} FUSIONCHAT-7B \cite{fusechat_wan2024} effectively leverages strengths from three architecturally diverse models (NH2-Mixtral-8x7B, NH2-Solar-10.7B, OpenChat-3.5-7B). Unlike traditional knowledge distillation, the fusion model surpasses all constituent models in reasoning (8.00) and coding (6.15), achieving competitive results against closed-source GPT-3.5 despite its modest 7B parameter size.

\textbf{Resource Requirements.} Knowledge fusion employs a two-stage process: knowledge transfer via lightweight fine-tuning followed by parameter-space merging using variation-ratio-based weights \cite{fusechat_wan2024}. With no size constraints on target models relative to sources, it requires modest computational resources, enabling creation of powerful composite models from diverse architectures without extensive pretraining datasets or prohibitive costs.

\textbf{Trade-offs.} While knowledge fusion creates systems exceeding individual source model performance, challenges remain in specialized areas where GPT-4 maintains advantages (8.55 vs 6.15 in coding). Recent advances in knowledge distillation techniques \cite{taid_shing2025} suggest promising pathways to further enhance efficiency by addressing capacity gaps through dynamic distribution interpolation. Despite being bounded by the collective knowledge of constituent models, this approach offers a promising direction for democratizing advanced language capabilities without relying on raw scaling.

\begin{table}[!htb]
\caption{Performance comparison of knowledge fusion approach on text and code generation tasks of mt-bench \cite{fusechat_wan2024}}
\tablefont%
\setlength{\tabcolsep}{2pt}
\begin{tabular*}{21pc}{@{} p{90pt} p{35pt}<{\raggedright\hangindent4pt} p{40pt}<{\raggedright\hangindent4pt} p{40pt}<{\raggedright\hangindent4pt} p{40pt}<{\raggedright\hangindent4pt} @{}}
\hline
Model & Type & Writing & Reasoning & Coding \\
\hline\\[-17pt]
&&&&\\

NH2-Mixtral-8x7B & 1 & \textbf{9.70} & 6.65 & 6.00 \\
NH2-Solar-10.7B & 2 & 9.50 & 7.35 & 4.00 \\
OpenChat-3.5-7B & 3 & 9.00 & 7.75 & 5.50 \\
FUSIONCHAT-7B VARM & 1 + 2 + 3 & 9.20 & \textbf{8.00} & 6.15 \\
GPT-3.5 (March) & -- & 9.20 & 5.65 & 6.90 \\

\hline
\multicolumn{5}{l}{}\\[-5pt]
\multicolumn{5}{@{}p{21pc}@{}}{\hspace*{9pt}FUSIONCHAT \cite{fusechat_wan2024} performs comparable to a closed-source GPT-3.5.}\\
\end{tabular*}
\label{tab:fusechat}
\end{table}

\subsubsection{Mixture-of-Experts} Mixture-of-Experts (MoE) architectures dynamically route inputs through specialized neural subnetworks, enabling efficient scaling of model capacity without proportional computational costs. Unlike traditional dense transformers, MoE models selectively activate only relevant experts for each input token. Recent implementations like Mixtral \cite{mixtralexperts_jiang2024} and OLMOE \cite{olmoe_muennighoff2025} demonstrate the effectiveness of this approach across diverse tasks (Table \ref{tab:MoE}).

\textbf{Key Findings.} Mixtral 8x7B \cite{mixtralexperts_jiang2024} achieves superior performance with significantly fewer active parameters, outperforming larger models like LLaMA 2 70B across multiple benchmarks. Despite using only 13B active parameters per token (from a total of 47B), it demonstrates stronger capabilities in MMLU (70.6\%) and code generation (MBPP: 60.7\%) compared to denser models. Similarly, OLMOE-1B-7B \cite{olmoe_muennighoff2025} shows remarkable efficiency, achieving state-of-the-art performance in code generation (HumanEval: 63.7\%) while using just 1B active parameters per token.

\textbf{Resource Requirements.} MoE architectures employ a router network at each layer to dynamically select relevant experts, typically activating only a small subset (e.g., 2 out of 8 experts in Mixtral) per token. This sparse activation pattern enables efficient inference despite large total parameter counts. Training requires specialized infrastructure to handle expert parallelism, but inference costs remain competitive due to the sparse computation pattern.

\textbf{Trade-offs.} While MoE models demonstrate superior parameter efficiency and performance across many tasks, they introduce additional complexity in routing mechanisms and expert load balancing. The approach shows particular promise in specialized tasks like code generation, where OLMOE matches or exceeds GPT-3.5's performance (63.7\% vs 62.8\% on HumanEval) with far fewer parameters. However, the need for specialized training infrastructure and careful expert design represents a notable implementation challenge compared to dense model.

\begin{table}[!htb]
\caption{MoE Performance on text and code generation}
\tablefont%
\setlength{\tabcolsep}{2pt}
\begin{tabular*}{21pc}{@{} p{90pt} p{45pt}<{\raggedright\hangindent4pt} p{45pt}<{\raggedright\hangindent4pt} p{45pt}<{\raggedright\hangindent4pt} <{\raggedright\hangindent4pt} @{}}
\hline
Model & MMLU & MBPP & HumanEval \\
\hline\\[-17pt]
&&&\\

LLaMA 2 70B & 69.9\% & 49.8\% & 29.3\%  \\
GPT-3.5-Turbo & 70.0\% & 52.2\% & 62.80\% \\
Mixtral 8x7B \cite{mixtralexperts_jiang2024} & \textbf{70.6\%} & \textbf{60.7\%} & 40.2\%  \\
OLMOE-1B-7B \cite{olmoe_muennighoff2025} & 56.3\% & -- & \textbf{63.7\%}  \\

\hline
\multicolumn{4}{l}{}\\[-5pt]
\multicolumn{4}{@{}p{21pc}@{}}{\hspace*{9pt}MoEs \cite{mixtralexperts_jiang2024, olmoe_muennighoff2025} with fewer parameters achieved significantly better functional
correctness than the use of a single LLM with a lot more parameters.}\\
\end{tabular*}
\label{tab:MoE}
\end{table}

\subsubsection{Reward Ensemble}

Reward ensemble methods enhance reinforcement learning from human feedback (RLHF) by aggregating multiple reward models to improve prediction accuracy and uncertainty estimation. Recent work in efficient reward ensembles \cite{rewardmodelensembleRlhf_zhang2024} and diverse LoRA approaches \cite{lowrankadoptation_zhai2023} demonstrates significant alignment improvements across benchmarks (Table \ref{tab:reward_ens}).

\textbf{Key Findings.} LoRA-based reward ensembles \cite{lowrankadoptation_zhai2023} outperformed both single reward models and simple ensemble averaging across metrics. They increased alignment scores from below 4.16 to 4.35 on MT-Bench with Llama-7B and improved accuracy from 0.685 to 0.710 on the Anthropic Helpful dataset. These improvements were achieved without proportional computational increases, effectively addressing the overoptimization problem common in RLHF.

\textbf{Resource Requirements.} These approaches maintain efficiency through lightweight techniques like linear-layer ensembles and LoRA-based methods. Diverse LoRA ensembles \cite{lowrankadoptation_zhai2023} maximize parameter efficiency by focusing adaptations on critical model components while encouraging diversity through nuclear norm maximization, requiring substantially fewer resources than multiple full reward models.

\textbf{Trade-offs.} While reward ensembles improve alignment, they introduce integration challenges for real-world applications. Limited evaluation on standardized benchmarks complicates adoption decisions, and production deployment requires complex orchestration of multiple reward models. Despite these challenges, the approach shows strong potential for resource-constrained environments, as reinforcement learning generally delivers superior performance when properly tuned, particularly if future work addresses evaluation gaps with comprehensive benchmarking against industry-standard metrics.

\begin{table}[!htb]
\caption{reward ensemble performance on text generation}
\tablefont%
\setlength{\tabcolsep}{2pt}
\begin{tabular*}{21pc}{@{} p{52pt} 
p{60pt}<{\raggedright\hangindent4pt} 
p{55pt}<{\raggedright\hangindent4pt} 
p{37pt}<{\raggedright\hangindent4pt} 
p{37pt}<{\raggedright\hangindent4pt} @{}}
\hline
Base Model & Dataset & Metric & Single RM & Ens. LoRA \\
\hline\\[-17pt]
&&&&\\

Llama-7B \cite{rewardmodelensembleRlhf_zhang2024} & MT-Bench & Alignment Score & \textless \ 4.16 &  4.35 \\
Llama2-7B \cite{lowrankadoptation_zhai2023} & Anthropic Helpful & Accuracy &  0.685 & 0.710 \\

\hline
\multicolumn{5}{l}{}\\[-5pt]
\multicolumn{5}{@{}p{21pc}@{}}{\hspace*{9pt}LoRA-based reward ensemble outperforms single-RM-based models.}\\
\end{tabular*}
\label{tab:reward_ens}
\end{table}

\subsubsection{Output Ensemble}

Output ensemble techniques combine responses from multiple language models to produce higher-quality outputs than individual models alone. Unlike weight merging or reward ensembles, output ensembles operate at inference time by aggregating multiple outputs. The Mixture-of-Agents (MoA) approach \cite{mixtureofagents_wang2024} demonstrates how layered agent architectures leverage collective model strengths without architectural modifications. The HumanEval results in Table \ref{tab:MoA} were obtained through our experiments combining four models (Llama-3.3-70B-Instruct-Turbo, Mixtral-8x7B-Instruct-v0.1, Qwen2.5-Coder-32B-Instruct, and DeepSeek-R1-Distill-Llama-70B-free) with Qwen2.5-Coder-32B-Instruct as the aggregator using Together API.

\textbf{Key Findings.} MoA \cite{mixtureofagents_wang2024} substantially outperforms GPT-4o mini across multiple domains. On AlpacaEval2.0, it achieves a 65.1\% win rate versus GPT-4o mini's 57.5\%. Similarly, it shows superior performance on alignment (MT-Bench: 9.25 vs 9.19) and code generation (HumanEval: 92.0\% vs 86.5\%). These improvements emerge from MoA's layered architecture, where each agent refines responses by considering outputs from previous layers before a final aggregator synthesizes a coherent response.

\textbf{Resource Requirements.} Output ensemble approaches require no model modification or retraining. MoA implements a model-level Mixture-of-Experts architecture where distinct LLMs function through prompt-based interactions. This architecture supports heterogeneous models regardless of size, architecture, or access level. As demonstrated in our experiments, combinations of open-source LLMs can match or exceed the performance of strong closed-source commercial models at potentially lower total cost, offering an economically viable alternative to expensive single-model solutions.

\textbf{Trade-offs.} While output ensembles improve performance across benchmarks, they introduce additional latency due to multiple model calls. The approach has proven versatile in applications from clinical text error detection \cite{promptmind_clinicaltext_gundabathula2024} and disease diagnosis \cite{votellms_oniani2024, diagnose_outputensemble_Barabucci2024} to image retrieval \cite{LDRE_yang2024} and instruction-following via a post-ranking fusion like LLM-Blender \cite{llmblender_jiang2023}. Despite increased inference complexity, output ensembles remain attractive for production environments where model updates are frequent, offering a practical path to high-quality language capabilities without expensive retraining.

\begin{table}[!htb]
\caption{output ensemble performance on text and code generation}
\tablefont%
\setlength{\tabcolsep}{3pt}
\begin{tabular*}{21pc}{@{} p{70pt} p{60pt}<{\raggedright\hangindent6pt} p{40pt}<{\raggedright\hangindent6pt} p{80pt}
<{\raggedright\hangindent6pt} @{}}
\hline
Model & AlpacaEval2.0 & MT-Bench  & HumanEval  \\
\hline\\[-17pt]
&&&\\
GPT-4o-mini (05/13) & 57.5\%   & 9.19  & 86.5\% \\
MoA & \textbf{65.1\%}  & \textbf{9.25}  & \textbf{92.0\%}  \\
\hline
\multicolumn{4}{l}{}\\[-5pt]
\multicolumn{4}{@{}p{21pc}@{}}{\hspace*{9pt}MoA-based open-source ensemble model \cite{mixtureofagents_wang2024} outperforms the commercial GPT-4o-mini in instruction-following and code generation tasks.}\\
\end{tabular*}
\label{tab:MoA}
\end{table}

\subsubsection{Routing}

Routing optimizes language model deployment by directing queries to the most suitable model from a pool of options, maximizing performance while minimizing costs. Recent advances \cite{harnessingpowermultipleminds_srivatsa2024, routoo_mohammadshahi2024} demonstrate remarkable efficiency gains across diverse tasks (Table \ref{tab:routing}).

\textbf{Key Findings.} Routoo \cite{routoo_mohammadshahi2024} shows that routing between open-source models can achieve 75.87\% accuracy on MMLU, surpassing GPT-3.5 at less than half the cost. When including closed-source models, it approaches the performance of GPT-4-Turbo (84. 9\% vs. 86. 4\%) at half the cost. The multilabel classifier (MLC) router \cite{harnessingpowermultipleminds_srivatsa2024} similarly outperforms single open-source models. RouterBench \cite{routerbench_hu2024} validates this approach in various domains, including code generation.

\textbf{Resource Requirements.} Routing systems use lightweight predictors to estimate model performance without full inference. Routoo implements a performance predictor and cost-aware selector with minimal overhead, enabling substantial cost savings while dynamically balancing quality, cost, and latency constraints based on deployment requirements.

\textbf{Trade-offs.} While routing systems offer impressive efficiency gains, their effectiveness depends on the prediction quality and the diversity of available models. Unlike ensembles that combine strengths across models, routing commits to a single model's response per query. Predictors may struggle with novel task distributions not in their training data. Despite these limitations, routing provides an immediate deployable solution for optimizing LLM deployments without ensemble complexity or model fusion demands, particularly valuable for applications with heterogeneous query types.

\begin{table}[!htb]
\caption{Routing Performance on text generation}
\tablefont%
\setlength{\tabcolsep}{2pt}
\begin{tabular*}{21pc}{@{} p{100pt} p{70pt}<{\raggedright\hangindent4pt} p{70pt}<{\raggedright\hangindent4pt} <{\raggedright\hangindent4pt} @{}}
\hline
Model & MMLU Acc. & Cost (\$/1M tok) \\
\hline\\[-17pt]
&&\\

LLaMA2-13B  &  54.8\% & 0.26  \\
Mistral-7B  & 62.09\% & 0.2  \\
MLC Router \cite{harnessingpowermultipleminds_srivatsa2024}  & 63.85\% & --  \\
GPT3.5  &  70.0\%  &  1.5  \\
Routoo (open-source) \cite{routoo_mohammadshahi2024}  &  75.87\%  &  0.6  \\
Routoo (mix) \cite{routoo_mohammadshahi2024}  &  84.9\%  &  10.2  \\
GPT4-Turbo &  86.40\%  &  20  \\

\hline
\multicolumn{3}{l}{}\\[-5pt]
\multicolumn{3}{@{}p{21pc}@{}}{\hspace*{9pt}The routing approach can significantly outperform single open-source LLMs \cite{harnessingpowermultipleminds_srivatsa2024}. Routing between open-source LLMs outperforms GPT-3.5 at less than half the cost \cite{routoo_mohammadshahi2024}. Mixing in GPT-4 achieves performance comparable to GPT-4-Turbo at half the cost \cite{routoo_mohammadshahi2024}.}\\
\end{tabular*}
\label{tab:routing}
\end{table}

\subsubsection{Cascading}
Cascading approaches deploy multiple LLMs in sequence, routing simpler queries to cost-effective models while reserving expensive models for complex tasks. Pioneered in FrugalGPT \cite{frugalgpt_chen2023} and enhanced with mixtures of thought representations \cite{llmcascades_yue2024}, cascades achieve near-optimal performance while significantly reducing computational costs.

\begin{figure*}[!htb]
  \centering
  \includegraphics[width=0.8\textwidth]{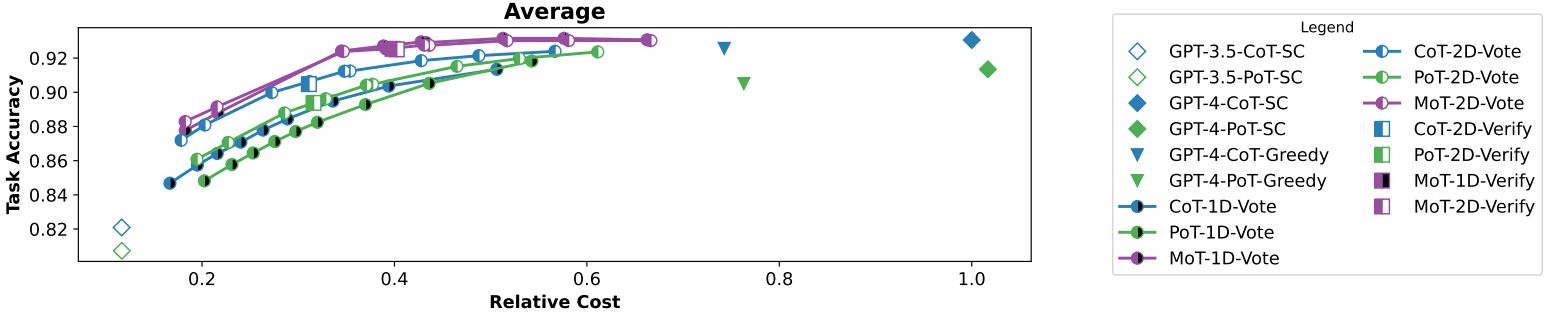}
  \caption{The performance curves of cost vs. accuracy of reasoning answer using (blue) the GPT with CoT models, (green) the GPT with PoT models, and (purple) the MoT cascade models \cite{llmcascades_yue2024}. The MoT cascade model performed similarly to the GPT-4 model with CoT prompts at half the cost in reasoning tasks.}
  \label{fig:llmcascade_plot}
\end{figure*}

\textbf{Key Findings.} FrugalGPT achieves 1.5\% higher accuracy than GPT-4 while reducing costs by 80\% on classification tasks (Table \ref{tab:cascading}). For reasoning tasks, models employing mixture of thought (MoT) representations match GPT-4's performance at approximately half the cost, using answer consistency from multiple reasoning paths as routing signals between weaker and stronger models (Figure \ref{fig:llmcascade_plot}).

\textbf{Resource Requirements.} Cascade systems require minimal overhead, primarily an efficient routing mechanism based on consistency checks \cite{llmcascades_yue2024} or learned selectors \cite{frugalgpt_chen2023}. This approach enables selective deployment of powerful models in cost-sensitive production environments.

\textbf{Trade-offs.} While cascades optimize costs, they introduce latency through sequential invocation and potential redundancy during query escalation. Their effectiveness depends on router accuracy—poor routing decisions can increase costs or degrade performance. Despite these challenges, cascading approaches demonstrate versatility across classification \cite{frugalgpt_chen2023} and reasoning tasks \cite{llmcascades_yue2024}, suggesting broader applicability in cost-constrained environments. Since coding requires reasoning skills, this cascade-based routing method could also generate optimal code at a lower cost.

\begin{table}[!htb]
\caption{cascading performance and cost on classification}
\tablefont%
\setlength{\tabcolsep}{3pt}
\begin{tabular*}{21pc}{@{} p{80pt} 
p{80pt}<{\raggedright\hangindent6pt} 
p{80pt}<{\raggedright\hangindent6pt} @{}}
\hline
Model & Accuracy & Cost (\$)  \\
\hline\\[-17pt]
&&\\
GPT-4 & 0.857   & 33.1  \\
FrugalGPT & \textbf{0.872}  & \textbf{6.5}  \\
\hline
\multicolumn{3}{l}{}\\[-5pt]
\multicolumn{3}{@{}p{21pc}@{}}{\hspace*{9pt}Cascading-based FrugalGPT \cite{frugalgpt_chen2023} reduces the cost by 80\%, while outperforms GPT-4 by 1.5\% in classification tasks on the HEADLINES dataset.}\\
\end{tabular*}
\label{tab:cascading}
\end{table}

\subsection{Comparative Analysis}
LLM ensemble approaches offer distinct advantages based on deployment constraints and optimization goals. Weight merging techniques provide training-free enhancements ideal for quick specialization without computational overhead but typically require architectural compatibility \cite{taskarithmetic,ties_merge_yadav2023,dare_yu2024}. Knowledge fusion accommodates structurally different models but demands more resources for fine-tuning \cite{knowledgefusionwan2024,fusechat_wan2024}. Mixture-of-Experts architectures excel in parameter efficiency by activating only relevant parameters, offering robust performance with controlled computational costs, though they require specialized pretraining \cite{mixtralexperts_jiang2024,olmoe_muennighoff2025} unlike output ensemble \cite{mixtureofagents_wang2024} and cascading approaches \cite{frugalgpt_chen2023,llmcascades_yue2024}.

The selection of ensemble method depends primarily on constraints: computational efficiency favors weight merging and routing; architectural flexibility benefits from knowledge fusion and output ensembles; production cost optimization is best served by cascading and routing approaches; and human preference alignment typically requires reward ensembles. As LLM capabilities evolve, hybrid approaches may emerge as the most practical path to democratizing advanced language capabilities across diverse deployment scenarios, combining strengths while mitigating individual weaknesses of each ensemble technique.

Table \ref{tab:allensemble} summarizes these recently proposed text and code generation approaches using LLM ensemble techniques. The model structure and advantage columns in the table also summarize different research directions. In the following "Discussion" section, we further analyze these approaches from three aspects based on their metrics, modeling traits and the comparison between text and code generations. 

\begin{table*}[!htb]
\caption{A summary of the proposed text and code generation approaches using LLM ensemble techniques}
\tablefont%
\setlength{\tabcolsep}{3pt}
\begin{tabular}{p{2.5cm} p{3.1cm} p{3.6cm} p{4.5cm} p{3.8cm}}
\hline
Ensemble Approach & Method & Model Structure & Advantage & Application \\ 
 \hline\\[-17pt]
&&\\

{\multirow{4}{*}{Weight Merging}}  
& Task Arithmetic \cite{taskarithmetic} \href{https://github.com/mlfoundations/task_vectors}{code} & {\multirow{3}{*}{Pretrained LLMs}} & {\multirow{3}{*}{Training-free performance enhancement}} & Math reasoning \\
{}  & TIES Merging \cite{ties_merge_yadav2023} \href{https://github.com/prateeky2806/ties-merging}{code} & {} & {} & Coding \\
{}  & TIES + DARE \cite{dare_yu2024} \href{https://github.com/yule-BUAA/MergeLM}{code} & {} & {} & Coding, instruction-following \\
{}  & EO \cite{GAMoEs_akiba2024} \href{https://github.com/SakanaAI/evolutionary-model-merge?tab=readme-ov-file}{code} & Evolve weights + merging & Automated model development & Math reasoning \\
\cmidrule(l){1-5}

{\multirow{2}{*}{Knowledge Fusion}}  & FuseLLM \cite{knowledgefusionwan2024} \href{https://github.com/fanqiwan/FuseAI}{code} & {\multirow{2}{*}{Fine-tuned LLMs}}  
& {\multirow{2}{*}{Can use structurally different models}} 
&  {\multirow{2}{*}{Reasoning, Coding}}   \\
{}  & FusionChat \cite{fusechat_wan2024} & {}
& {}
& {} \\
\cmidrule(l){1-5}

{\multirow{2}{*}{Mixture-of-Experts}}  & Mixtral \cite{mixtralexperts_jiang2024} \href{https://github.com/mistralai/mistral-inference}{code} & {\multirow{2}{*}{Pretrained transformer}} 
& {\multirow{2}{*}{Parameter efficient robust performance}} &  {\multirow{2}{*}{MMLU, Coding}} \\
{}  & OLMOE \cite{olmoe_muennighoff2025} \href{https://github.com/allenai/OLMoE}{code} & {} & {} & {} \\
\cmidrule(l){1-5}

{\multirow{2}{*}{Reward Ensemble}}  & RLHF + LoRA \cite{rewardmodelensembleRlhf_zhang2024} & {\multirow{2}{*}{Fine-tuned LLMs + RL}}  
& {\multirow{2}{*}{Superior in human alignment}} & {\multirow{2}{*}{Conversation}} \\
{}  & RLHF + LoRA \cite{lowrankadoptation_zhai2023} & {} & {} & {} \\
\cmidrule(l){1-5}

{\multirow{2}{*}{Output Ensemble}}  & MoA \cite{mixtureofagents_wang2024} \href{https://github.com/togethercomputer/moa}{code} & Feed-forward LLMs 
& Training-free agentic workflow &  Instruction-following, Coding \\
{}  & LLM Blender \cite{llmblender_jiang2023} \href{https://github.com/yuchenlin/LLM-Blender}{code} & Ranker and fuser transformers & Extension flexible component & Instruction-following \\
\cmidrule(l){1-5}

{\multirow{2}{*}{Routing}}  & MLC Router \cite{harnessingpowermultipleminds_srivatsa2024} \href{https://github.com/kvadityasrivatsa/llm-routing}{code} & Multi-label classifier 
& {\multirow{2}{*}{Cost effective without complexity}} &  {\multirow{2}{*}{MMLU}}  \\
{}  & Routoo \cite{routoo_mohammadshahi2024} \href{https://github.com/Leeroo-AI/leeroo_orchestrator}{code} & 2 lightweight LLMs & {} & {} \\
\cmidrule(l){1-5}

{\multirow{2}{*}{Cascading}}  & FrugalGPT \cite{frugalgpt_chen2023} \href{https://github.com/stanford-futuredata/FrugalGPT}{code} & Sequence of LLMs 
& {\multirow{2}{*}{Cost effective with broader applications}} &  Classification \\
{}  & MoT \cite{llmcascades_yue2024} \href{https://github.com/MurongYue/LLM_MoT_cascade}{code} & MoT + Sequence of LLMs & {} & Math reasoning \\
\cmidrule(l){1-5}

\multicolumn{3}{l}{}\\[-5pt]
\multicolumn{3}{@{}p{21pc}@{}}{\hspace*{9pt}Model Structure and Advantage can summarize various research directions.}\\
\end{tabular}
\label{tab:allensemble}
\end{table*}

\section{Discussion}

\subsection{Metrics}


Metrics vary by task and model type. As discussed in Section \ref{sec:background}, LLMs excel at capturing long-range dependencies in language features that correlate with human perception. This capability informs evaluation approaches for both text and code generation across LLM ensemble methods.
For text generation, human-perceptual correlations can be measured using metrics like BERTScore \cite{bertscore_zhang2020} or BARTScore \cite{bartscore_yuan2021}, which compare generated output to ground truth text \cite{llmblender_jiang2023}. Similarly, code generation quality can be evaluated using CodeBERTScore \cite{codebertscore_zhou2023}, which applies the same underlying principles to code snippets.
In cascade and routing models, predicted quality scores determine which LLM's response is selected as final output \cite{llmcascades_yue2024, frugalgpt_chen2023, routerbench_hu2024, routoo_mohammadshahi2024}, making response quality a direct indicator of routing effectiveness.

\subsection{Model Structures and Model Training}
Table \ref{tab:allensemble} shows that different model structures are used in these seven approaches to achieve various goals. MoE employs pretrained transformers for gating and expert functions, allowing for input-to-expert routing \cite{mixtralexperts_jiang2024,olmoe_muennighoff2025}. Similarly, weight merging uses pretrained LLMs to interpolate/extrapolate weights, enabling training-free performance enhancement \cite{taskarithmetic, ties_merge_yadav2023, dare_yu2024}. 
Building on those pretrained models, knowledge fusion fine-tune them while fusing structurally different models \cite{knowledgefusionwan2024, fusechat_wan2024}. 
In contrast, reward ensemble uses those fine-tuned LLMs as backbones and combine them with RL approach for improved human alignment \cite{rewardmodelensembleRlhf_zhang2024,lowrankadoptation_zhai2023}.

Given the high computational cost of tuning LLMs and other transformer models, one option is to skip the training process altogether by using a model-level output ensemble.
As a training-free agentic workflow, output ensemble models structure LLMs in a feed-forward way, enhancing performance comparable to commercial LLMs \cite{mixtureofagents_wang2024}. Alternatively, transformers can still be trained for multioutput ranking and fusion \cite{llmblender_jiang2023}. Both cases deal with LLM endpoints only.

In routing-based models, routers help invoke multiple LLMs sequentially \cite{frugalgpt_chen2023,llmcascades_yue2024} or route to the best single LLM \cite{harnessingpowermultipleminds_srivatsa2024, routoo_mohammadshahi2024}. They aim to produce cost-efficient outputs and use various scoring strategies. These include agreement scores to measure LLM consistency for reasoning \cite{llmcascades_yue2024}, reward scores for instruction-following \cite{rewardrouter_lu2023}, and scoring methods using classifier, transformers, or LLMs for inference optimization \cite{harnessingpowermultipleminds_srivatsa2024,hybridllm_ding2024,routoo_mohammadshahi2024}. 



\subsection{A Comparison Between Text and Code Generations Both Using LLM Ensemble Models}
While LLM ensemble architectures like weight-merging, knowledge fusion, MoE, and MoA share model structures across text and code generation \cite{dare_yu2024,knowledgefusionwan2024,fusechat_wan2024,mixtralexperts_jiang2024,olmoe_muennighoff2025,mixtureofagents_wang2024} (Table \ref{tab:allensemble}), their optimal application and the role of individual experts within the ensemble diverge significantly. These differences stem primarily from the nature of the generation target and the consequent evaluation paradigms.

A core distinction lies in the characteristics of the output. Text generation grapples with natural language's inherent ambiguity, stylistic variety, and broad knowledge domains. Ensembles in this space often aim to synthesize diverse reasoning capabilities, blend stylistic nuances, or enhance robustness in complex instruction following. In contrast, code generation targets formal, structured programming languages where syntactic and semantic correctness are paramount for execution. Thus, ensemble models for code might specialize in ensuring logical integrity, generating specific valid constructs, or even different programming paradigms, with a stronger emphasis on verifiable correctness over subjective stylistic qualities.

These differing objectives are directly reflected in evaluation methodologies, which in turn influence ensemble design. For text, metrics often focus on instruction-following accuracy (e.g., AlpacaEval win rates), answer correctness (e.g., MMLU accuracy), and linguistic quality (e.g., BERTScore \cite{bertscore_zhang2020}), potentially leading to aggregation strategies that enhance fluency or consensus. Code generation, however, prioritizes functional correctness (e.g., pass@k on benchmarks like HumanEval), runtime, and memory usage. This frequently drives ensemble designs towards generating a diverse set of candidate solutions to maximize the probability of at least one passing unit tests, or selecting the best verifiable output. Metrics adapted from text generation, such as CodeBERTScore \cite{codebertscore_zhou2023} and CodeBLEU \cite{codebleu_ren2020}, evaluate the similarity of generated code against a reference solution, capturing aspects like semantic equivalence even without exact token matches. However, this assessment of closeness to an exemplar provides a different signal than the direct confirmation of executional correctness and functional behavior offered by definitive pass/fail unit tests. Recognizing these distinct pressures is crucial for effectively adapting ensemble techniques from one domain to the other.

\section{Conclusion}

This review highlights recent LLM ensemble learning techniques for language generation. Methods such as weight merging, MoE, MoA, and routing enhance LLM diversity and output combination, improving the quality of generated text and code compared to single-model approaches. We systematically analyzed representative implementations for text and code generation to provide a clear overview of their real-world applicability.

Key advantages identified include:

1) Weight merging boosts performance without additional training.
2) Knowledge fusion enables fine-tuning of structurally different models.
3) Parameter-efficient MoE models can outperform much larger LLMs.
4) Training-free MoA models offer easy integration and task adaptability.
5) Routing and cascading provide cost-effective scalability for powerful LLMs like GPT-4 and Gemini.
These findings are relevant for model selection in diverse applications like chatbot assistance, medical diagnosis, mathematical reasoning, and code optimization.

Despite these successes, challenges remain. Current weight merging and knowledge fusion methods show limitations in coding performance, suggesting a need for ensemble structures specifically designed for programming languages.

Furthermore, a significant frontier is extending ensemble principles to multimodal LLMs. While this review focused on text and code, the core concept of leveraging diverse model strengths is highly applicable to vision-language models like BLIP-2, Flamingo, and LLaVA. Future research should actively investigate ensembling multimodal outputs (e.g., via ranking and fusion), intelligent routing, knowledge fusion/weight merging across multimodal architectures, and specialized multimodal MoEs. Advancing ensemble strategies for both specialized code generation and the evolving landscape of multimodal AI is crucial for unlocking the full potential of collaborative intelligence in a wider range of applications.





\bibliography{bibtex/bib/IEEEabrv.bib,bibtex/bib/IEEEexample.bib}{}
\bibliographystyle{IEEEtranN}

\end{document}